\documentclass[10pt]{article} 
\usepackage[accepted]{rlc}

\usepackage{amssymb}            
\usepackage{mathtools}          
\usepackage{mathrsfs}           
\usepackage{graphicx}           
\usepackage{subcaption}         
\usepackage[space]{grffile}     
\usepackage{url}                

\def \mcT{\mathcal{T}}

\def \mcS{\mathcal{S}}

\def \mcA{\mathcal{A}}

\def \ev{\mathbb{E}}

\newcommand{\vo}{\vec{o}\@ifnextchar{^}{\,}{}}

\newtheorem{innercustomgeneric}{\customgenericname}
\providecommand{\customgenericname}{}
\newcommand{\newcustomtheorem}[2]{%
  \newenvironment{#1}[1]
  {%
   \renewcommand\customgenericname{#2}%
   \renewcommand\theinnercustomgeneric{##1}%
   \innercustomgeneric
  }
  {\endinnercustomgeneric}
}

\newcustomtheorem{customthm}{Theorem}
\newcustomtheorem{customcrlry}{Corollary}
\newcustomtheorem{customlemma}{Lemma}
\newcustomtheorem{customassumption}{Assumption}

\usepackage{multirow}

\title{Online Statistical Inference of constant Sample-averaged Q-Learning}

\author{Saunak Kumar Panda  \\
    spanda@uh.edu \\
    Department of Industrial Engineering\\
    University of Houston
    \And
    Tong Li \\
    tli31@cougarnet.uh.edu\\
    Department of Industrial Engineering\\
    University of Houston
    \And
    Ruiqi Liu \\
    ruiqliu@ttu.edu\\
    Department of Mathematics and Statistics\\
    Texas Tech University
    \And
    Yisha Xiang \\
    yxiang4@central.uh.edu\\
    Department of Industrial Engineering\\
    University of Houston}

\begin{document}

\maketitle

\begin{abstract}
Reinforcement learning algorithms have been widely used for decision-making tasks in various domains. However, the performance of these algorithms can be impacted by high variance and instability, particularly in environments with noise or sparse rewards. In this paper, we propose a framework to perform statistical online inference for a sample-averaged Q-learning approach. We adapt the functional central limit theorem (FCLT) for the modified algorithm under some general conditions and then construct confidence intervals for the Q-values via random scaling. We conduct experiments to perform inference on both the modified approach and its traditional counterpart, Q-learning using random scaling and report their coverage rates and confidence interval widths on two problems: a grid world problem as a simple toy example and a dynamic resource-matching problem as a real-world example for comparison between the two solution approaches.
\end{abstract}

\section{Introduction}
\label{sec:introduction}

Reinforcement learning (RL) has emerged as a powerful paradigm for training agents to make sequential decisions in complex and uncertain environments. While many RL algorithms have shown remarkable success in a variety of applications, there is a growing recognition of the need to incorporate statistical inference techniques into RL to enhance its robustness, reliability, and interpretability. This motivation stems from the inherently stochastic nature of RL environments, where randomness can arise from various sources, such as sensor noise, unobservable state variables, or even the inherent uncertainty in the agent's interactions with the environment. Statistical inference plays an important role in several real-world applications ranging from financial analysis of stocks and other financial assets to critical applications like medical research where confidence intervals of mean difference in outcomes between treatment and control groups are calculated to determine whether a new drug or treatment is effective. Several frequently employed methods in statistical inference, particularly for computing confidence intervals, encompass bootstrapping, spectral variance, and batch-means, among others. The majority of these methods leverage the central limit theorem (CLT) to attain the asymptotic covariance of estimate distributions of our RL algorithms. One promising research direction is the application of the functional central limit theorem (FCLT) to RL algorithms, particularly in the context of Q-learning. The FCLT, which provides a theoretical foundation for the emergence of normality in certain functionals of random processes, offers a novel perspective for understanding the statistical properties of Q-learning and related algorithms. Leveraging FCLT results in the context of Q-learning can help provide insights into its convergence behavior and uncertainty quantification, making it a valuable tool for characterizing its statistical properties. This paper aims to explore and leverage the FCLT to advance our understanding of statistical inference, specifically for a sample-averaged variant of the Q-learning algorithm, called sample-averaged Q-learning. Our main contribution lies in providing theoretical guarantees for our sample-averaged Q-learning algorithm under standard assumptions and further demonstrate higher accuracy of our method using a random scaling approach for computing confidence measures on a few numerical examples. 

The remainder of the paper is organized as follows. In Section 2, we review relevant literature on statistical inference methods, in the context of stochastic gradient descent(SGD) and RL. In Section 3, we provide the problem setup and propose our sample-averaged Q-learning. In Section 4, we discuss the main theoretical result in detail and provide the method to perform statistical inference for the proposed sample-averaged Q-learning. In Section 6, we present numerical results and compare the confidence measures of Q-values obtained from both the vanilla Q-learning method and our proposed sample-averaged Q-learning approach on two numerical examples. Our approach showcases higher accuracy in terms of the coverage rates and confidence interval lengths.

\section{Literature Review}
The literature on statistical inference methods in the context of SGD and RL has seen significant growth in recent years, with researchers proposing various methodologies to address the challenges associated with making statistical inferences from data generated by these methods, specifically iid and Markovian data. 

\citet{li2018statistical} introduce a statistical inference framework using SGD. Their methodology relies on the asymptotic properties of SGD and established convergence results. It provides a foundation for making statistical inferences on parameters learned through SGD optimization. This approach is less expensive than bootstrapping for large data size and parameter dimensionality, that is, there is no need for re-sampling many times. An added advantage is that this approach only uses first-order information and we can also conduct finite sample inference. \citet{fang2018online} present online bootstrap confidence intervals for the SGD estimator. By incorporating bootstrapping techniques into the analysis, they proposed a method for constructing confidence intervals for the parameters learned through SGD, addressing the problem of estimating uncertainty in this context. \citet{chen2020statistical} propose statistical inference for model parameters in SGD. Their work provides a comprehensive framework involving two methods, namely batch-means and plug-in estimators for quantifying uncertainty in parameter estimates. They show that the plug-in estimator achieves faster convergence than batch means, while the batch-means estimator needs less storage and calculation. \citet{chen2021statistical} propose a statistical inference method for online decision-making via SGD. Their approach utilizes asymptotic normality results to construct confidence intervals for online decision rules. They derived an inverse probability-weighted value estimator to estimate the optimal value, and proposed plug-in estimators to estimate the variance of the parameters. \citet{lee2022fast} propose a novel technique through an FCLT result that enables them to develop an efficient online inference method via random scaling. They show the robustness of their inference method for SGD algorithms using simulation experiments. The above methods are well suited to cases where the data is generated by an iid sampling procedure and are unable to account for the underlying dependence structure of the observations following a Markovian structure.

\citet{ramprasad2023online} propose an online bootstrap inference method for policy evaluation in RL. They leverage the bootstrap technique to construct confidence intervals for policy value estimates in both on- and off-policy estimates. The approach adapts to non-i.i.d. and temporally correlated data encountered in RL tasks, making it suitable for dynamic environments. However, this method may be computationally intensive and sensitive to the choice of bootstrap parameters. \citet{li2023statistical} propose the Polyak-Ruppert-Averaged Q-learning algorithm and provide an FCLT result showing their algorithm is statistically efficient. However, they require stronger conditions on their noise moments. \citet{xie2022statistical} introduce a statistical online inference approach in averaged stochastic approximation. They develop a unified framework for analyzing algorithms such as SGD and Q-learning from the perspective of stochastic approximation. They establish an FCLT for time-homogeneous Markov chains under weaker conditions and then construct confidence intervals for parameters via random scaling. Since the random scaling approach uses a studentized approach to obtaining the variance, it avoids additional estimation steps and hyper-parameters, resulting in its higher accuracy over traditional estimators like bootstrapping and batch-means.

\section{Problem Setup}
An infinite-horizon MDP is represented by $\mathcal{M}=(\mathcal{S}, \mathcal{A}, \gamma, P, R, r)$, where $\mathcal{S}$ is the state space and $\mathcal{A}$ is the action space, $\gamma \in(0,1)$ is the discount factor, $P: \mathcal{S} \times \mathcal{A} \rightarrow \Delta(\mathcal{S})$ represents the probability transition kernel, $R: \mathcal{S} \times \mathcal{A} \rightarrow[0, \infty)$ is the random reward, and $r: \mathcal{S} \times \mathcal{A} \rightarrow[0, \infty)$ is the expectation of the reward. For a given deterministic policy $\pi: \mathcal{S} \rightarrow \mathcal{A}$, the Q-function for any state-action pair $(s,a) \in \mcS \times \mcA$ is given by,
\begin{align}
    Q^\pi(s,a) &= \ev_{\tau \sim \pi} \left[\sum_{t=0}^{\infty} \gamma^t r(s_t,a_t) | s_0 = s,a_0 =a \right], \label{eq:Q-function}
\end{align}
where $\tau = \{(s_t,a_t)\}_{t \geq0}$ is a trajectory of the MDP induced by the policy $\pi$. The optimal Q-function $Q^*$ is defined as $Q^*(s, a)=\max _\pi Q^\pi(s, a)$. We know that $Q^*$ is the unique fixed point of the Bellman operator $\mathcal{T}$, where
$$
\mathcal{T}(Q)(s, a)=r(s, a)+\gamma \mathbb{E}_{s^{\prime} \sim P(\cdot \mid s, a)} \max _{a^{\prime} \in \mathcal{A}} Q\left(s^{\prime}, a^{\prime}\right) .
$$
Suppose that we have a sequence of i.i.d. observations $\{(r_i,s'_i),i=1,2,\dotso\}$ given state action pair $(s, a)$. For each $(s,a) \in \mcS \times \mcA$ at the $t$-step, we generate $B_t$ rewards $R_{t,1}(s,a),\ldots, R_{t, B_t}(s,a)$ and the associate next states $S'_{t,1}, \ldots, S'_{t,B_t}$. We consider dividing them into several batches.

To estimate $Q^*$, the sample-averaged Q-learning will have the following update rule:
\begin{align}\label{eq:BQL}
Q_{t+1}(s,a)&=Q_t(s,a)- \eta_t\left\{Q_t(s,a)-\widehat{\mcT}_{t+1}(Q_t)(s,a)\right\}\\
\end{align}
where 
\begin{align}
 \widehat{\mathcal{T}}_{t+1}(Q_t)(s,a)=\frac{1}{B_t}\sum_{i=1}^{B_t} R_{t,i}(s,a)+ \frac{\gamma}{B_t}\sum_{i=1}^{B_t} \max_{a_i'}Q_t(S'_{t,i}, a_i').
\end{align}
and $\gamma$ is the discount factor. In particular, when $B_t=1$ is a constant, it reduces to vanilla Q-learning. Hence, our sample-averaged approach can be considered as a generalization of the single-sample vanilla Q-learning algorithm. It can be easily shown that $\widehat{\mcT}_t$ is an unbiased estimator of the Bellman operator $\mcT$.

\section{Asymptotic Results}
In this section, we provide a theoretical result for the constant sample-averaged Q-learning update given in equation (\ref{eq:BQL}) such that $B_t=B \geq 1$, where $B$ is a constant. Specifically, to state our FCLT result, we first show that our sample-averaged Q-learning process converges to a stationary distribution and, we further provide an upper bound for the bias arising from the gap between the expectation and the ground truth solution. Finally, we establish the FCLT for our sample-averaged Q-learning algorithm which enables us to perform inference using a random scaling method. 

\begin{customassumption}{A1}\label{assumption:A1}\ 
(Uniformly bounded reward). The random reward $R$ is non-negative and uniformly bounded, i.e., $0 \leqslant R(s, a) \leqslant 1$ almost surely for all $(s, a) \in \mcS \times \mcA$.
\end{customassumption}
In the above assumption, we impose a boundedness condition on the random reward of the model.

\subsection{Functional Central Limit Theorem for sample-averaged Q-learning}

\begin{customthm}{1}\label{thm:1}
    Assume access to a generative model for each state-action pair $(s, a) \in \mcS \times \mcA$. Let $\textbf{Q} \in \mathbb{R}^{|\mcS|\times|\mcA|}$ be the matrix form of the Q-function $Q(s, a)$. Then there exists a constant $\eta_0 > 0$ such that for any $\eta \in (0, \eta_0)$,

    \begin{enumerate}
        \item the sample-averaged Markov process $(\mathbf{Q}_t)_{t \geq 0}$ defined by (\ref{eq:BQL}) has a unique stationary distribution $\mathbf{Q}_\eta$
        \item for $\mathbf{Q} \sim \mathbf{Q}_\eta$, we have $\mathbb{E}_{\mathbf{Q}_\eta} ||\mathbf{Q} -\mathbf{Q}^*||_\infty = \mathcal{O}(\eta^{1/2})$
        \item The following FCLT holds
        \begin{align}
            \frac{1}{\sqrt{\sum_{t=1}^T B_t^{-1}}} \sum_{t=1}^{\lfloor rT \rfloor} (\mathbf{Q}_t - \mathbb{E}_{\mathbf{Q}_\eta}\mathbf{Q}) \Rightarrow \Sigma_{\mathbf{Q}_\eta}^{\frac{1}{2}}M(r), \qquad 0 \leq r \leq 1,
        \end{align}
        where  $(M_r))_{r \geq 0}$ is the d-dimensional standard Brownian motion and $\Sigma_{Q_\eta} = \lim_{n \xrightarrow{} \infty} n\text{cov}_{Q_\eta}(\Bar{Q}_n)$ is the asymptotic covariance matrix.
    \end{enumerate}
\end{customthm}
A direct consequence of Theorem \ref{thm:1} with $r=1$ is the asymptotic distribution of the sample-averaged estimator via averaging, which is summarized as the following corollary.
\begin{customcrlry}{1}
Under the uniform bounded reward assumption, it follows that
\begin{align*}
\frac{T}{\sqrt{\sum_{t=1}^T B_t^{-1}}}\left(\bar{\mathbf{Q}}_T-\mathbf{Q}^*\right) \xrightarrow{\mathbb{L}} N(0, \Omega)    
\end{align*}
\end{customcrlry}
Noting that there are total $N=\sum_{t=1}^T B_t$ observations used after $T$ iterations, by Corollary 1 , we can quantify the asymptotic covariance of $\bar{\mathbf{Q}}_T$ by
$$
\operatorname{Var}\left(\sqrt{N} \bar{\mathbf{Q}}_T\right) \approx N \frac{\sum_{t=1}^T B_t^{-1}}{T^2} \Omega=\frac{\sum_{t=1}^T B_t \sum_{t=1}^T B_t^{-1}}{T^2} \Omega .
$$

It can be verified that the estimator $\bar{\mathbf{Q}}_T$ with a constant batch size $B_t=B$ is asymptotically efficient with an asymptotic covariance matrix $\Omega$. However, when $B_t$ is diverging, it suffers from an efficiency loss. Obtaining an efficient sample-averaged estimator for Q-learning is out of the scope of this paper.

\subsection{Online inference via Random Scaling}

In this section, we propose an online algorithm based on Theorem \ref{thm:1} for constructing a confidence interval of $\mathbf{Q}^*$. Before proceeding, let us introduce the random scaling quantity:
$$
\widehat{D}_T=\frac{1}{T} \sum_{s=1}^T\left\{\frac{1}{m_T} \sum_{t=1}^s\left(\mathbf{Q}_t-\bar{\mathbf{Q}}_T\right)\right\}\left\{\frac{1}{m_T} \sum_{t=1}^s\left(\mathbf{Q}_t-\bar{\mathbf{Q}}_T\right)\right\}^{\top},
$$
where $m_T=\sqrt{\sum_{t=1}^T B_t^{-1}}$. If we further define a stochastic process
$$
\widehat{M}(r):=m_T^{-1} \sum_{t=1}^{\lceil r T\rceil}\left(\mathbf{Q}_t-\mathbf{Q}^*\right)
$$
it can be verified that $\widehat{M}(1)=T m_T^{-1}\left(\bar{\mathbf{Q}}_T-\mathbf{Q}^*\right)$ and
$$
\widehat{D}_T=\int_0^1(\widehat{M}(r)-r \widehat{M}(1))(\widehat{M}(r)-r \widehat{M}(1))^{\top} d r .
$$

As a consequence of the continuous mapping theorem and Theorem \ref{thm:1}, we obtain the following result, which can be easily shown using Theorem \ref{thm:1} and continuous mapping theorem.

\begin{customthm}{2}\label{thm:2}
Under our bounded reward assumption, it holds that
$$
\frac{T^2}{m_T^2}\left(\bar{\mathbf{Q}}_T-\mathbf{Q}^*\right) \widehat{D}_T^{-1}\left(\bar{\mathbf{Q}}_T-\mathbf{Q}^*\right)^{\top} \xrightarrow{\mathbb{L}} M(1)\left\{\int_0^1 \bar{M}(r) \bar{M}^{\top}(r) d r\right\}^{-1} M^{\top}(1) .
$$

Here $M(r)$ is d-dimensional standard Brownian motion on $[0,1]$, and $\bar{M}(r)=M(r)-$ $r M(1)$. As a consequence, it follows that
$$
\widehat{\kappa}=\frac{T\left(\bar{\mathbf{Q}}_{T, j}-\mathbf{Q}_j^*\right)}{m_T \sqrt{\widehat{D}_{T, j j}}} \xrightarrow{\mathbb{L}} M_1(1)\left(\int_0^1\left\{M_1(r)-r M_1(1)\right\}^2 d r\right)^{-1 / 2}:=\kappa .
$$

Here $\bar{\mathbf{Q}}_{T, j}, \mathbf{Q}_j^*$ are the $j$-th entry of $\bar{\mathbf{Q}}_T, \mathbf{Q}^*, \widehat{D}_{T, j j}$ is the $j$-th diagonal element of $\widehat{D}_T$, and $M_1(r)$ is a one-dimensional standard Brownian motion.    
\end{customthm}

Theorem \ref{thm:2} shows that we can construct a statistic $\widehat{\kappa}$ that is asymptotically pivotal. As a consequence, we can construct a $(1-\alpha) \times 100 \%$-level confidence interval for $\mathbf{Q}_j^*$ as follows:
$$
\bar{\mathbf{Q}}_{T, j} \pm \kappa_{\alpha / 2} \frac{m_T}{T} \sqrt{\widehat{D}_{T, j j}} .
$$

Here $\kappa_{\alpha / 2}$ is the upper $\alpha / 2$-quantile of the random variable $\kappa$. As shown in \citet{ca5c88cc-60e5-392c-ad0a-5a051aa7c51f}, the distribution of $\kappa$ is a mixed normal and symmetric around zero. The values of $\kappa_\alpha$ can be found in Table 1 in \citet{ca5c88cc-60e5-392c-ad0a-5a051aa7c51f}.

\section{Numerical experiments}
In this section, we will conduct a computational study to assess the performance of our proposed sample-averaged Q-learning algorithm as compared to the vanilla Q-learning algorithm using the random scaling method. We will first use a grid world problem to illustrate the performance in a small state and action space. We will further provide a case study in the context of real-world dynamic supply-demand matching, where the states and corresponding actions are high-dimensional and their spaces are significantly larger.

\subsection{Grid World Problem}
\begin{figure}[htp]
    \centering
    \includegraphics{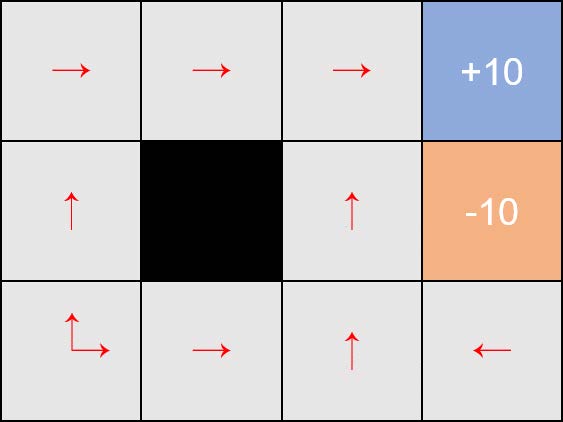}
    \caption{Grid World}
    \label{fig:fig1}
\end{figure}

In this section, we consider synchronous Q-learning in a Grid World environment. The environment consists of 3 × 4 grids as its state space. At each grid, the agent can choose to go “up”, “down”, “left” or “right” into the next grid. If it touches the edge or the block (i.e., the black grid), then it stays still. It stops after arriving at the blue/orange grid. Each immediate reward is -1 at grey grids, and +10/ - 10 at the blue/orange grid, respectively. The discount factor is set to $\gamma$ = 0.9. Under this deterministic reward setting, the optimal policy at each grid is represented by a red arrow in Figure~\ref{fig:fig1}. 

Now, we consider randomizing the immediate reward at each grid with a $\mathcal{N}(0, \sigma^2)$ Gaussian noise. We set $\sigma$ = 2 and the learning rate $\eta$ = 0.1. The performance of the random scaling method for Vanilla Q-learning and sample-averaged Q-learning (batch) are compared in three aspects: the coverage
rates for $\mathbb{E}_{Q_\eta}\mathbf{Q}$ and $\mathbf{Q}^*$, and the lengths of confidence intervals. Here, the expected Q-value $\mathbb{E}_{\mathbf{Q}_\eta}\mathbf{Q}$ is approximated by 500,000 Monte Carlo simulations, and the true Q-value $\mathbf{Q}^*$ is derived by Q-learning in the deterministic reward setting. The nominal coverage rate is chosen as 95\%. We perform 10,000 iterations of both algorithms on this problem.

It can be observed from Table~\ref{table:tab1} that the coverage rates for the random scaling (RS) method for vanilla Q-learning are consistently higher than the ones with sample-averaged Q-learning. It can be seen that the confidence interval lengths converge to approximately the same values for both cases. However, to draw more conclusions about both the sampling Q-learning methods, examples with more complexity need to be considered as the current grid world example has few states and action spaces and, therefore is not extensive enough to draw conclusive results. Hence, we next consider a non-trivial learning problem.

\begin{table}[htbp]
\centering
\caption{Statistical Inference for Q-learning for Grid World Problem}
\begin{tabular}{ccccc}
\hline
                                                      & \multicolumn{1}{c|}{}  & \multicolumn{1}{c|}{$n$ = 2000} & \multicolumn{1}{c|}{$n$ = 6000} & \multicolumn{1}{c|}{$n$ = 10000} \\ \hline
$\eta=0.1, \sigma=2$                                  & \multicolumn{4}{c}{}                                                                                                         \\ \cline{1-1}
\multirow{2}{*}{Cov. for $\mathbf{Q}^*$ (\%)} & RS for Vanilla QL      & 99.0 (0.054)                            & 97.5 (0.021)                           & 96.0 (0.0153)                            \\
                                                      & RS for Sample-averaged QL & 99.0 (0.075)                           & 99.0 (0.025)                           & 99.0 (0.0158) 
                                                      \\ \hline
\multirow{2}{*}{Cov. for $\mathbb{E}_{\mathbf{Q}_\eta}\mathbf{Q}$ (\%)} & RS for Vanilla QL      & 99.0 (0.054)                           & 98.0 (0.021)                           & 97.5  (0.0153)                          \\
                                                      & RS for Sample-averaged QL & 99.0 (0.075)                           & 99.0 (0.025)                           & 99.0 (0.0158)                            \\ \hline
\multirow{2}{*}{Length}                               & RS for Vanilla QL      & 0.735                          & 0.285                          & 0.21                          \\
                                                      & RS for Sample-averaged QL & 1.0                          & 0.345                           & 0.215                           \\ \hline
\end{tabular}
\label{table:tab1}
\end{table}

\subsection{Dynamic Matching Problem}

\begin{figure}[htp]
    \centering
    \includegraphics[width=0.7\linewidth]{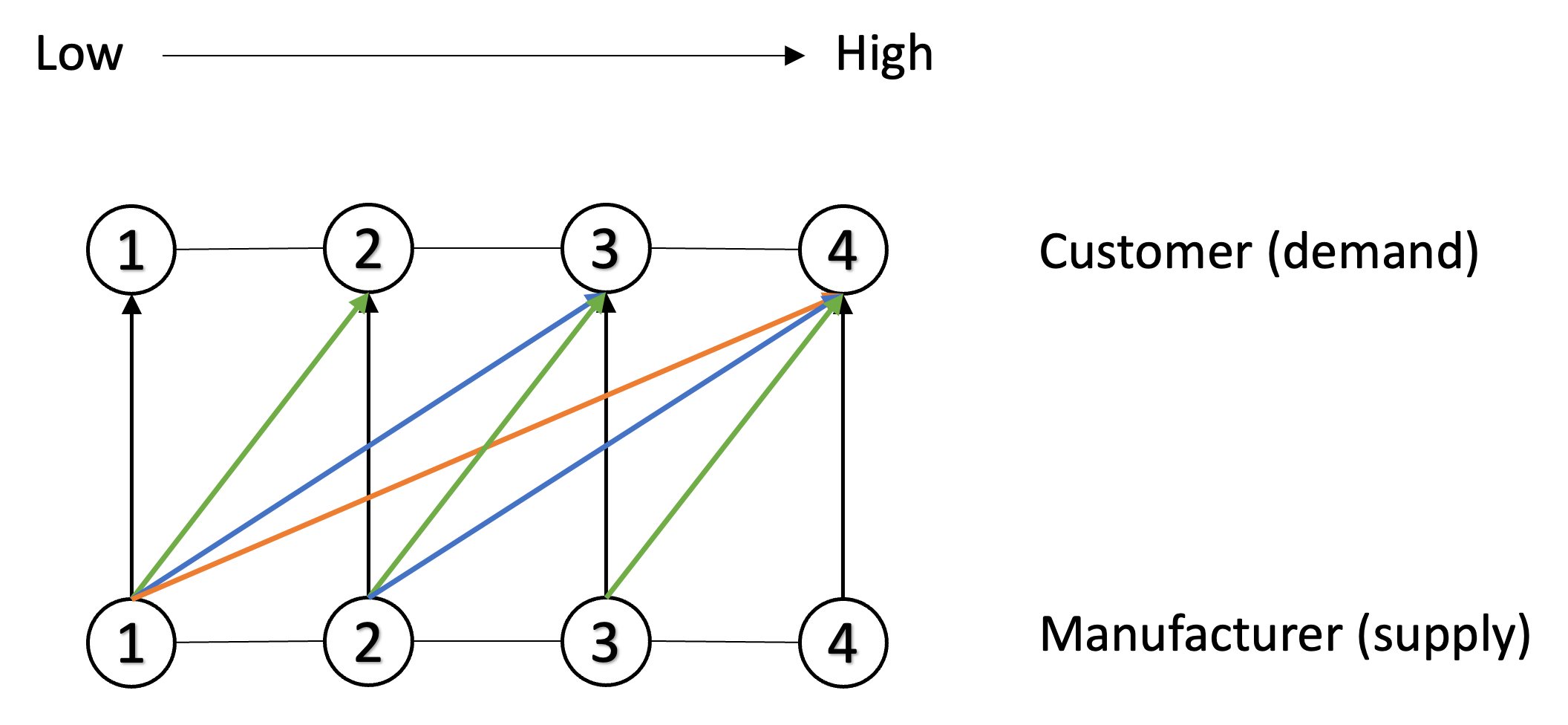}
    \caption{Dynamic Resource-matching}
    \label{fig:fig2}
\end{figure}
We consider Q-learning in a dynamic resource-matching problem represented in Figure\ref{fig:fig2}. For simplicity, the system consists of 2 demand and 2 supply types, that is, a 2x2 matching problem. The reward matrix $\textbf{R}$ is chosen such that $r_{11}>r_{21}>r_{12}>r_{22}$ and is given by \(\textbf{R} = [8 \quad5\quad;\quad7\quad3]\). Each immediate reward is the product of the quantity matched with the reward matrix. The possible demand quantities for both types are $(0,1,2,3)$ with probabilities $(0.5,0.5,0,0)$ for both type 1 and type 2 products. The discount factor is set to $\gamma$ = 0.9.

We consider randomizing the immediate reward for each matching with a $\mathcal{N}(0, \sigma^2)$ Gaussian noise. We set $\sigma$ = 2 and the learning rate $\eta$ = 0.1. The performance of the random scaling method for Vanilla Q-learning and sample-averaged Q-learning (batch) with a sample size $B=5$ are compared in three aspects: the coverage
rates for $\mathbb{E}_{\mathbf{Q}_\eta}\mathbf{Q}$ and $\mathbf{Q}^*$, and the lengths of confidence intervals. Here, the expected Q-value $\mathbb{E}_{\mathbf{Q}_\eta}\mathbf{Q}$ is approximated by 5,000 Monte Carlo simulations, and the true Q-value $\mathbf{Q}^*$ is derived by Q-learning in the deterministic reward setting. The nominal coverage rate is chosen as 95\%. Since this problem contains larger state and action spaces, we perform 2,000 iterations of both algorithms on this problem.

It can be observed from Table~\ref{table:tab2} that while the coverage rates for the random scaling (RS) method for both the Q-learning methods are approximately the same, the confidence interval lengths are much tighter for the sample-average case, indicating much better accuracy in the confidence measures for the dynamic resource-matching problem. 

\begin{table}[htbp]
\centering
\caption{Statistical Inference for Q-learning for Dynamic Matching Problem}
\begin{tabular}{ccccc}
\hline
                                                      & \multicolumn{1}{c|}{}  & \multicolumn{1}{c|}{$n$ = 500} & \multicolumn{1}{c|}{$n$ = 1000} & \multicolumn{1}{c|}{$n$ = 2000} \\ \hline
$\eta=0.1, \sigma=2$                                  & \multicolumn{4}{c}{}                                                                                                         \\ \cline{1-1}
\multirow{2}{*}{Cov. for $\mathbf{Q}^*$ (\%)} & RS for Vanilla QL      & 99.9                            & 99.9                            & 99.9                            \\
                                                      & RS for Sample-averaged QL & 99.9                            & 99.9                            & 99.9                            \\ \hline
\multirow{2}{*}{Cov. for $\mathbb{E}_{\mathbf{Q}_\eta}\mathbf{Q}$ (\%)} & RS for Vanilla QL      & 99.9                            & 99.9                            & 99.9                           \\
                                                      & RS for Sample-averaged QL & 99.9                            & 99.9                            & 99.9                            \\ \hline
\multirow{2}{*}{Length}                               & RS for Vanilla QL      & 299.4                          & 199.5                           & 113.8                           \\
                                                      & RS for Sample-averaged QL & 49.6                          & 33.7                           & 19.1                           \\ \hline
\end{tabular}
\label{table:tab2}
\end{table}

\section{Conclusion}
In this paper, we proposed a statistical inference procedure for a sample-averaged variant of Q-learning. Based on an FCLT in the context of Markov chains, we established a corresponding FCLT result for our proposed sample-averaged Q-learning. We conducted numerical experiments for both the Q-learning variants and showed higher accuracy for our sample-averaged Q-learning approach using random scaling. Some interesting extensions can be developing a random scaling method for the following variants: (i) an adaptive sample-averaged Q-learning algorithm where the batch size $B_t$ changes adaptively with time; (ii) a linear function approximation (LFA) of the proposed sample-averaged Q-learning. Extending the above approach to scheduled batch and LFA requires considering additional rate conditions and assumptions on the linearly independent functions representing the state, which can be regarded as future work.

\section*{Acknowledgements}
This work is supported in part by the U.S. National Science Foundation under award 2305486.

\appendix
\section*{Appendix}
\section{Proof of Theorem~\ref{thm:1}}\label{sec:appendixA}

We will first verify that $\widehat{\mathcal{T}}_{t}$ is an unbiased estimator of the Bellman operator $\mathcal{T}$. By direct examination, (\ref{eq:BQL}) can be written as
    \begin{align}\label{eq:BQL2}
       Q_{t+1}(s,a)=Q_t(s,a)+\eta_t \underbrace{\left\{\widehat{\mathcal{T}}_{t+1}(Q_t)(s,a)-Q_t(s, a)\right\}}_{\hat{W}_{t+1}(Q_t)(s,a)}.
    \end{align}
Now,
\begin{align*}
    \mathbb{E} \Large \{ \widehat{W}_{t+1}(Q_t)(s,a)|Q_t(s,a) \Large \} &= \mathbb{E} \{ \widehat{\mathcal{T}}_{t+1}(Q)(s,a) - Q_t(s,a) \}\\
                                                                    &= \mathbb{E} \{ \frac{1}{B_t} \sum_{i=1}^{B_t} r_{t,i}(s,a) + \gamma \frac{1}{B_t} \sum_{i=1}^{B_t} \max_{a'_i} Q(s_{t,i},a'_i) - Q_t(s,a) |Q_t(s,a)\}
                                                                    \intertext{By linearity of expectation,}
                                                                    &= \frac{1}{B_t} \mathbb{E} \{  \sum_{i=1}^{B_t} r_{t,i}(s,a) |Q_t(s,a)\} + \frac{\gamma}{B_t} \mathbb{E}\{\sum_{i=1}^{B_t} \max_{a'_i} Q(s_{t,i},a'_i)|Q_t(s,a)\}\\
                                                                    &\qquad -\mathbb{E} \{Q_t(s,a) |Q_t(s,a)\}\\
                                                                    &= \frac{1}{B_t} \sum_{i=1}^{B_t}\mathbb{E} \{   r_{t,i}(s,a) |Q_t(s,a)\} + \frac{\gamma}{B_t} \sum_{i=1}^{B_t} \mathbb{E}\{ \max_{a'_i} Q(s_{t,i},a'_i) |Q_t(s,a) \} \\
                                                                    &\qquad - Q_t(s,a)
                                                                    \intertext{Since the data is collected from a generative model,}
                                                                    &= r_t(s,a) + \gamma \mathbb{E}_{s' \sim P(\cdot|s,a)} \{\max_{a'}Q(s',a')\} - Q_t(s,a)
                                                                    \intertext{Using the definition of $\mathcal{T}(Q)(s,a)$,}
                                                                    &= \mathcal{T}(Q)(s,a) - Q_t(s,a)
\end{align*}
Hence, from RHS we proved that $\widehat{\mathcal{T}}_{t}$ is an unbiased estimator of $\mathcal{T}$.\\

Now, denote $\mathcal{L}(X)$ as the distribution of the R.V. X. We need to prove that $\{\mathcal{L}(Q_t)\}_{t \geq 0}$ is a Cauchy sequence in the complete space $\mathcal{P}_{\infty,1}$.\\

Suppose $Q_0 \sim \mu$, and take any positive integer $N>0$. For any $k \geq N$ and $l \geq 0$, we need to show that $\mathcal{W}_{\infty,1}\large (\mathcal{L}(Q_k),\mathcal{L}(Q_{k+l}) \large)$ is bounded, where for $p, q \geq 1$, the Wasserstein distance $\mathcal{W}_{p, q}$ induced from the $L^p\left(\mathbb{R}^d\right)$ space is defined by
$$
\mathcal{W}_{p, q}(\mu, \nu)=\inf _{\gamma \in \Gamma(\mu, \nu)}\left(\int_{\mathbb{R}^d \times \mathbb{R}^d}\|x-y\|_p^q \mathrm{~d} \gamma(x, y)\right)^{\frac{1}{q}},
$$
where $\Gamma(\mu, \nu)$ is the set of all couplings of $\mu$ and $\nu$; and denote $\mathcal{P}_{p, q}$ as its corresponding Wasserstein space.\\

Consider the process (\ref{eq:BQL}) started from $Q_0^{(1)} \sim \mathcal{L}(Q_0)$ and $Q_0^{(2)} \sim \mathcal{L}(Q_l)$. In the first $k$ iterations, we couple the two processes with the same randomness. From Lemma B.3 in \citet{xie2022statistical},

\begin{align*}
    \mathcal{W}_{\infty,1}\large (\mathcal{L}(Q_k),\mathcal{L}(Q_{k+l}) \large) &= \mathcal{W}_{\infty,1}\large (\mathcal{L}(Q_k^{(1)}),\mathcal{L}(Q_k^{(2)}) \large)\\
                                                                                &\leq (1-\eta - \eta \gamma)^k \mathcal{W}_{\infty,1}\large (\mathcal{L}(Q_0^{(1)}),\mathcal{L}(Q_0^{(2)}) \large)\\
                                                                                & \leq (1-\eta - \eta \gamma)^N \mathbb{E}||Q_l-Q_0||_\infty.
\end{align*}
Since $0 \leq r_t(s,a) \leq 1$, for any $l \geq 0$,

\begin{align*}
    \mathbb{E}||Q_{l+1}||_\infty &= \mathbb{E}||(1-\eta)Q_l + \eta \widehat{\mathcal{T}}_{l+1}(Q_l)||_\infty \\
                                 &\leq \mathbb{E} \{||(1-\eta)Q_l||_\infty + ||\eta \widehat{\mathcal{T}}_{l+1}(Q_l)||_\infty \}
                                 \intertext{By linearity of expectation,}
                                 &= (1-\eta)\mathbb{E}||(Q_l||_\infty + \eta \mathbb{E} ||\widehat{\mathcal{T}}_{l+1}(Q_l)||_\infty
                                 \intertext{Since the reward variable is bounded,}
                                 &\leq (1-\eta)\mathbb{E}||(Q_l||_\infty + \eta + \eta \mathbb{E} \{\frac{\gamma}{B} \sum_{i=1}^B \max_{a'_i} Q_l(s_{l+1,i},a'_i)\}
                                 \intertext{By linearity of expectation,}
                                 &\leq (1-\eta)\mathbb{E}||(Q_l||_\infty + \eta + \frac{\eta \gamma}{B} \sum_{i=1}^B\mathbb{E} \{\max_{a'_i} Q_l(s_{l+1,i},a'_i)\}
                                 \intertext{Since the data is collected from a generative model,}
                                 &= (1-\eta)\mathbb{E}||(Q_l||_\infty + \eta + \eta \gamma \mathbb{E}_{s' \sim P(\cdot|s,a)} \{\max_{a'} Q_l(s_{l+1},a')\}
                                 \intertext{By the definition of $\infty$-norm,}
                                 &= (1-\eta)\mathbb{E}||(Q_l||_\infty + \eta + \eta \gamma \mathbb{E}||\max_{a'} Q_l(s_{l+1},a')||_\infty\\
                                 &= (1 - \eta + \eta \gamma) \mathbb{E}||Q_l||_\infty + \eta.
\end{align*}

By induction, we have for any $l \geq 0$,
\begin{align*}
    \mathbb{E}||Q_l||_\infty &\leq (1-\eta -\eta \gamma)^l E||Q_0||_\infty + \frac{\eta}{1 - (1-eta + \eta \gamma)}\\
                             &=(1-\eta -\eta \gamma)^l E||Q_0||_\infty + \frac{1}{1-\gamma},
\end{align*}
and thus, $\mathbb{E}||Q_l||_\infty$ is uniformly bounded for all $l \geq 0$. As $N \xrightarrow{} \infty$, we have $\mathcal{W}_{\infty,1}\large (\mathcal{L}(Q_k),\mathcal{L}(Q_{k+l}) \large) \xrightarrow{} 0$, and therefore $\{\mathcal{L}(Q_t)\}_{t \geq 0}$ is a Cauchy sequence in $\mathcal{P}_{\infty,1}$, and its limit $\mathcal{Q}_\eta$ exists. Uniqueness can be easily proven using Lemma B.3 from \citet{xie2022statistical}. Furthermore, the remaining statements 2. and 3. can be verified similarly as provided in \citet{xie2022statistical}.

\bibliography{main}
\bibliographystyle{rlc}

\end{document}